\documentclass[11pt,a4paper]{article}
\usepackage[hyperref]{acl2019}
\usepackage{times}
\usepackage{latexsym}
\usepackage{times}  
\usepackage{helvet}  
\usepackage{courier}  
\usepackage{url}  
\usepackage{graphicx}  
\usepackage[inline]{enumitem}
\usepackage{graphicx}
\usepackage{subfig}
\usepackage{float}
\usepackage[protrusion=true,expansion=true]{microtype}
\usepackage{graphics}
\usepackage{breqn}
\usepackage{amssymb}
\usepackage{caption,booktabs}

\usepackage{url}

\aclfinalcopy 

\title{Towards Coherent and Engaging Spoken Dialog Response Generation Using Automatic Conversation Evaluators}

\author{Sanghyun Yi$^1$, Rahul Goel$^2$, Chandra Khatri$^3$, Alessandra Cervone$^4$, Tagyoung Chung$^5$, \\{\bf Behnam Hedayatnia$^5$, Anu Venkatesh$^5$,  Raefer Gabriel$^5$, Dilek Hakkani-Tur$^5$}\\
  $^1$Division of Humanities and Social Sciences, California Institute of Technology,\\
  $^2$Google,
   $^3$Uber AI, 
   $^5$Alexa AI, Amazon\\
   $^4$Signals and Interactive Systems Lab, University of Trento\\
  {\tt syi@caltech.edu, goelrahul@google.com},\\
  {\tt \{tagyoung,behnam,anuvenk,raeferg,hakkanit\}@amazon.com},\\
  {\tt chdrak@uber.com}, 
  {\tt alessandra.cervone@unitn.it}}

\date{}

\begin{document}
\maketitle
\begin{abstract}

Encoder-decoder based neural architectures serve as the basis of
state-of-the-art approaches in end-to-end open domain dialog systems.
Since most of such systems are trained with a maximum likelihood~(MLE) objective they suffer from issues such as lack of generalizability and the {\it generic
response problem}, i.e., a system response that can be an answer to a large number of user utterances, e.g., ``Maybe, I don't know.'' Having explicit feedback on the relevance and interestingness of a system response at each turn
can be a useful signal for mitigating such issues and improving system quality by selecting responses from different approaches.  Towards this goal, we present a system that evaluates chatbot responses at each dialog turn
for coherence and engagement. Our system provides explicit turn-level dialog
quality feedback, which we show to be highly correlated with human evaluation.
To show that incorporating this feedback in the neural response generation
models improves dialog quality, we present two different and complementary
mechanisms to incorporate explicit feedback into a neural response generation
model: reranking and direct modification of the loss function during training.
Our studies show that a response generation model that incorporates these
combined feedback mechanisms produce more engaging and coherent responses in an
open-domain spoken dialog setting, significantly improving the response quality
using both automatic and human evaluation.


\end{abstract}

\section{Introduction}\label{sec:intro}

Due to recent advances in spoken language understanding and automatic speech
recognition, conversational interfaces such as Alexa, Cortana, and Siri have become increasingly common.  While these interfaces are task oriented, there is an increasing interest in building conversational systems that
can engage in more social conversations. Building systems that can have a general conversation in an open domain setting is a challenging problem, but it is an important step towards more natural human-machine interactions. 

Recently, there has been significant interest in building
chatbots~\cite{sordoni2015neural,wen2015semantically}
fueled by the availability of dialog data sets such as Ubuntu, Twitter, and Movie dialogs~\cite{lowe2015ubuntu,ritter2011data,Danescu-Niculescu-Mizil+Lee:11a}.
However, as most chatbots are text-based, work on human-machine spoken dialog is relatively under-explored, partly due to lack of such dialog corpora. 
Spoken dialog poses additional challenges such as automatic speech recognition
errors and divergence between spoken and written language. 

Sequence-to-sequence (seq2seq) models \cite{sutskever2014sequence} and their
extensions~\cite{luong2015effective,sordoni2015neural,li2015diversity}, which
are used for neural machine translation (MT), have been widely adopted for
dialog generation systems.  In MT, given a source sentence, the correctness of
the target sentence can be measured by semantic similarity to the source
sentence.  However, in open-domain conversations, a generic utterance such as
``sounds good'' could be a valid response to a large variety of statements.
These seq2seq models are commonly trained on a maximum likelihood objective,
which leads the models to place uniform importance on all user utterance and
system response pairs. Thus, these models usually choose ``safe'' responses as
they frequently appear in the dialog training data. This phenomenon is known as
the {\it generic response problem.} These responses, while arguably correct, are
bland and convey little information leading to short conversations and low user
satisfaction.

Since response generation systems are trained by maximizing the average
likelihood of the training data, they do not have a clear signal on how well the
current conversation is going. We hypothesize that having a way to measure
conversational success at every turn could be valuable information that can
guide system response generation and help improving system quality. Such a measurement may also be useful for combining responses from various competing
systems. To this end, we build a supervised conversational evaluator to assess
two aspects of responses: engagement and coherence. The input to our evaluators are encoded conversations represented as fixed-length vectors as well as hand-crafted dialog and turn level features. 
The system outputs explicit scores
on coherence and engagement of the system response.

We experiment with two ways to incorporate these explicit signals in response generation systems. First, we use the evaluator outputs as input to a reranking model, which are used to rescore the $n$-best outputs obtained after beam search
decoding.  Second, we propose a technique to incorporate the evaluator loss
directly into the conversational model as an additional discriminatory loss
term. Using both human and automatic evaluations, we show that both of these
methods significantly improve the system response quality. The combined model
utilizing re-ranking and the composite loss outperforms models using either
mechanism alone.

The contributions of this work are two-fold. First, we experiment with various
hand-crafted features and conversational encoding schemes to build a conversational evaluation system that can provide explicit turn-level feedback to a response generation system on the highly subjective task. This system can
be used independently to compare various response generation systems or as a
signal to improve response generation. Second, we experiment with two
complementary ways to incorporate explicit feedback to the response generation
systems and show improvement in dialog quality using automatic metrics as well
as human evaluation.

\section{Related Works}
There are two major themes in this work. The first is building evaluators that
allow us to estimate human perceptions of coherence, topicality, and
interestingness of responses in a conversational context.  The second is the use
of evaluators to guide the generation process. As a result, this work is related
to two distinct bodies of work.

\textbf{Automatic Evaluation of Conversations}: 
Learning automatic evaluation of conversation quality has a long
history~\cite{walker1997paradise}. However, we still do not have widely accepted
solutions. Due to the similarity between conversational response generation and
MT, automatic MT metrics such as BLEU~\cite{papineni2002bleu} and
METEOR~\cite{banerjee2005meteor} are widely adopted for evaluating dialog
generation. ROUGE~\cite{lin2003automatic}, which is also used for chatbot
evaluation, is a popular metric for text summarization.  These metrics primarily
rely on token-level overlap over a corpus (also synonymy in the case of METEOR),
and therefore are not well-suited for dialog generation since a valid
conversational response may not have any token-level or even semantic-level
overlap with the ground truths. While the shortcomings of these metrics are well
known for
MT~\cite{graham2015accurate,espinosa2010further}, the
problem is aggravated for dialog generation evaluation because of the much
larger output space~\cite{liu2016not, novikova2017we}. However, due to the lack of clear
alternatives, these metrics are still widely used for evaluating response
generation~\cite{ritter2011data,lowe2017towards}. To ensure comparability with
other approaches, we report results on these metrics for our models.


To tackle the shortcomings of automatic metrics, there have been  efforts to
build models to score conversations. \citet{lowe2017towards} train a model to
predict the score of a system response given a dialog context.
However, they work
with tiny data sets (around 4000 sentences) in a non-spoken
setting. \citet{tao2017ruber}  address the expensive annotation process
by adding in unsupervised data. However, their metric is not
interpretable, and the results are also not shown on a spoken setting. Our work
differs from the aforementioned works as the output of our system is interpretable
at each dialog turn.


There has also been work on building evaluation systems that focus on specific
aspects of dialog. \citet{li2016deep} use features for information flow,
\citet{yu2016strategy} use features for turn-level appropriateness.
However, these
metrics are based on a narrow aspect of the conversation and fail to capture
broad ranges of phenomena that lead to a good dialog.

\textbf{Improving System Response Generation}: Seq2Seq models have allowed
researchers to train dialog models without relying on handcrafted dialog acts
and slot values.
Using maximum mutual information (MMI)~\cite{li2015diversity} was one of the
earlier attempts to make conversational responses more diverse~\cite{serban2016building,vhred2016}. \citet{shao2017generating} use a segment ranking beam search to produce more
diverse responses. Our method extends the strategy employed
by~\citet{shao2017generating} utilizing a trained model as the reranking
function and is similar to ~\citet{holtzman2018learning} but with different kind of trained model. 


More recently, there have been works which aim to alleviate this problem by
incorporating conversation-specific rewards in the learning
process. \citet{yao2016attentional} use the IDF value of generated sentences as
a reward signal. \citet{xing2017topic} use topics as an additional input
while decoding to produce more specific responses. \citet{li2016persona}
add personal information to make system responses more user
specific.\citet{li2017data} use distillation to train different models at
different levels of specificity and use reinforcement learning to pick the
appropriate system response. \citet{zhou2017mechanism}
and \citet{zhang2018learning} introduce latent factors in the seq2seq
models that control specificity in neural response generation.  There has been
recent work which combines responses from multiple
sub-systems \cite{serban2017deep, papaioannou2017alana} and ranks them to output
the final system response. Our method complements these approaches by
introducing a novel learned-estimator model as the additional reward signal.

\section{Data} \label{sec:data}
The data used in this study was collected during the Alexa Prize~\cite{ram2017}
competition and shared with the teams who were participating in the competition.
Upon initiating the conversation, users were paired with a randomly selected
chatbot built by the participants. At the end of the conversation, the users
were prompted to rate the chatbot quality, from 1--5, with 5 being the highest.


We randomly sampled more than 15K conversations (approximately 160K
turns) collected during the competition. These were annotated for coherence and
engagement (See Section~\ref{subsec:Annotations}) and used to train the
conversation evaluators. For training the response generators, we selected
highly-rated user conversations, which resulted in around 370K conversations
containing 4M user utterances and their corresponding system response. One
notable statistic is that user utterances are typically very short (mean:
3.6 tokens) while the system responses generally are much longer (mean:
23.2 tokens).




\subsection{Annotations} \label{subsec:Annotations}
Asking annotators to measure coherence and engagement directly is a time-consuming task. We observed that we could collect data much faster if we asked
direct ``yes'' or ``no'' questions to our annotators. Hence, upon reviewing a
user-chatbot interaction along with the entire conversation to the current turn,
annotators\footnote{The data was collected through mechanical turk. Annotators
were presented with the full context of the dialog up to the current turn.} rated
each chatbot response as ``yes'' or ``no'' on the following criteria:
\begin{itemize}  [ topsep=3pt, partopsep=1pt, parsep=2pt]
  \setlength{\itemsep}{1pt}
\item \textbf{The system response is comprehensible:} The information provided
  by the chatbot made sense with respect to the user utterance and is syntactially correct.
\item \textbf{The system response is on topic:} The chatbot response
  was on the same topic as the user utterance or was relevant to the user utterance. For example, if a user asks about a baseball player on the LA
Dodgers, then the chatbot mentions something about the baseball
  team.
\item \textbf{The system response is interesting:} The chatbot response contains information which is novel and relevant. For example, the chatbot would
  provide an answer about a baseball player and give some additional information to create a fleshed-out response.
\item \textbf{I want to continue the conversation:} Given the current state of the conversation and the system response, there is a natural way to continue
  the conversation. For example, this could be due to the system asking a
  question about the current conversation subject.
\end{itemize}

 We use these questions as proxies for measuring coherence and engagement of responses. The answers to the first two questions (``comprehensible'' and ``on topic'') are used as a proxy for coherence. Similarly, the answer to the last two questions
 (``interesting'' and ``continue the conversation'') are used as a
proxy for engagement.

\section{Conversation Evaluators} \label{sec:eval}

We train conversational response evaluators to assess the state of a given
conversation. Our models are trained on a combination of utterance and response
pairs combined with context (past turn user utterances and system responses)
along with other features, e.g., dialog acts and topics as described in
Section~\ref{ssec:features}. We experiment with different ways to encode the
responses (Section \ref{sec:embd}) as well as with different feature combinations
(Figure \ref{fig:eval_model}).

\begin{table}
\centering
\scalebox{0.85}{
\begin{tabular}{l|l|l|l}
\bf Model & \bf TREC & \bf SUBJ  & \bf STS \\
\hline
\bf Average Emb. & 0.80 & 0.90  & 0.45 \\
 \bf Transformer & 0.83 & 0.91 & 0.48  \\
 \bf BiLSTM & 0.84 & 0.90 &  0.45  \\
 \end{tabular}
}
\caption { Sentence embedding performance.} \label{tab:sent_emb}
\end{table}

\subsection{Sentence Embeddings} \label{sec:embd}

We pretrained models that produce sentence embeddings using
the ParlAI chitchat data set~\cite{miller2017parlai}. We use the Quick-Thought~(QT)
loss~\cite{logeswaran2018efficient} to train the embeddings.
Our word embeddings are initialized with FastText~\cite{bojanowski2016enriching}
to capture the sub-word features and then fine-tuned. We encode sentences into
embeddings using the following methods:
\begin{enumerate} [label=\itshape\alph*\upshape), topsep=4pt, partopsep=2pt, parsep=4pt]
\setlength{\itemsep}{0pt}
\item Average of word embeddings (300 dim)
\item The Transformer Network~(1 layer, 600 dim)~\cite{vaswani2017attention}
\item Concatenated last states of a BiLSTM~(1 layer, 600 dim)
\end {enumerate}
The selected dimensions and network structures followed the original paper~\cite{vaswani2017attention}. 
All models were trained with a batch size of 400 using Adam optimizer with learning rate of 5e-4\@.


To measure the sentence embedding quality, we evaluate our models on a few
standard classification tasks. 
The models are used to get sentence representation, which are passed through feedforward networks that are trained for the following classification tasks: (i)
Semantic Textual Similarity (STS)~\cite{marelli2014sick}, (ii) Question Type
Classification (TREC)~\cite{voorhees2003overview}, (iii) Subjectivity
Classification (SUBJ)~\cite{pang2004sentimental}. Table~\ref{tab:sent_emb} shows
the different models' performances on these tasks. Based on this, we choose
the Transformer as our sentence encoder as it was overall the best performing
while being fast.


\subsection{Context}
Given the contextual nature of the problem
  we extracted the sentence embeddings of user utterances and responses for the past 5 turns and
  used a 1 layer LSTM with 256 hidden units to encode conversational context. The last state
  of LSTM is used to obtain the encoded representation, which is then
  concatenated with other features (Section \ref{ssec:features}) in a fully-connected neural network.
  
\subsection{Features} \label{ssec:features}
Apart from sentence embeddings and context, the following features are also used:

\begin{itemize} [ topsep=4pt, partopsep=2pt, parsep=4pt]
\setlength{\itemsep}{0pt}
\item \textbf{Dialog Act:} \citet{serban2017deep} show that dialog act~(DA)
  features could be useful for response selection rankers. Following this, we use model~\cite{khatri2018contextual}-predicted DAs~\cite{stolcke1998dialog} of user utterances and system responses as an indicator feature.
\item \textbf{Entity Grid:} \citet{cervone2018coherence, barzilay2008modeling} show that entities and
  DA transitions across turns can be strong features for assessing dialog coherence. Starting from a grid representation of the turns of the conversation as a matrix (DAs $\times$ entities), these features are designed to capture the patterns of topic and intent shift distribution of a dialog. We employ the same strategy for our models.
\item \textbf{Named Entity~(NE) Overlap:} We use named entity overlap between
  user utterances and their corresponding system responses as a feature. Our
  named entities are obtained using SpaCy\footnote{https://spacy.io/}.
  \citet{papaioannou2017alana} have also used similar NE features in their
  ranker.
\item \textbf{Topic:} We use a one-hot representation of a dialog turn 
  topic predicted by a conversational topic
  model~\cite{guo2017evaluating} that classifies a given dialog turn into one of
   26 pre-defined classes like Sports and Movies.
\item \textbf{Response Similarity:} Cosine similarity between user utterance
  embedding and system response embedding is used as a feature.
\item \textbf{Length:} We use the token-level length of the user utterance and
  the response as a feature.
\end{itemize}
The above features were selected from a large pool of features through
significance testing on our development set. 
The effect of adding these features can be seen in Table~\ref{tab:eval_results}. 
Some of the features such as {\bf Topic} lack previous dialog context, which could be updated to include the
context. We leave this extension for future work.

\begin{table*}[h]
\centering
\scalebox{0.85}{
\begin{tabular}{c|c|c|c|c|c|c}
\bf Evaluator & \bf {`Yes' Class Distr.} & \bf Accuracy & \bf Precision  & \bf Recall & \bf F-score & \bf MCC \\
\hline
Comprehensible & 0.80 & 0.84 (+3\%)  & 0.83 (+1\%) & 0.85 (+15\%)  & 0.84 (+8\%) &  0.37 (+107\%)\\
On-topic & 0.45 & 0.64 (+9\%) & 0.65 (+10\%) & 0.64 (+18\%) & 0.64 (+13\%) & 0.29 (+81\%)\\
Interesting & 0.16 & 0.83 (-1\%) & 0.77 (+10\%) & 0.80 (-5\%) & 0.78 (+2\%) & 0.12 (+inf\%)\\
Cont. Conversation & 0.71 & 0.75 (+4\%) & 0.73 (+5\%) & 0.72 (+31\%) & 0.72 (+17\%) & 0.32(+179\%)\\
\end{tabular}
} \caption { Conversation Evaluators Performance. Numbers in parentheses
  denote relative changes when using our best model (all features)
  with respect to the baseline (no handcrafted features, only sentence embeddings).
  Second column shows the class imbalance in our annotations. 
  Note that the baseline
  model had 0 MCC for Interesting} \label{tab:eval_results} \end{table*}

\begin{figure}[t]
  \includegraphics[width=\linewidth]{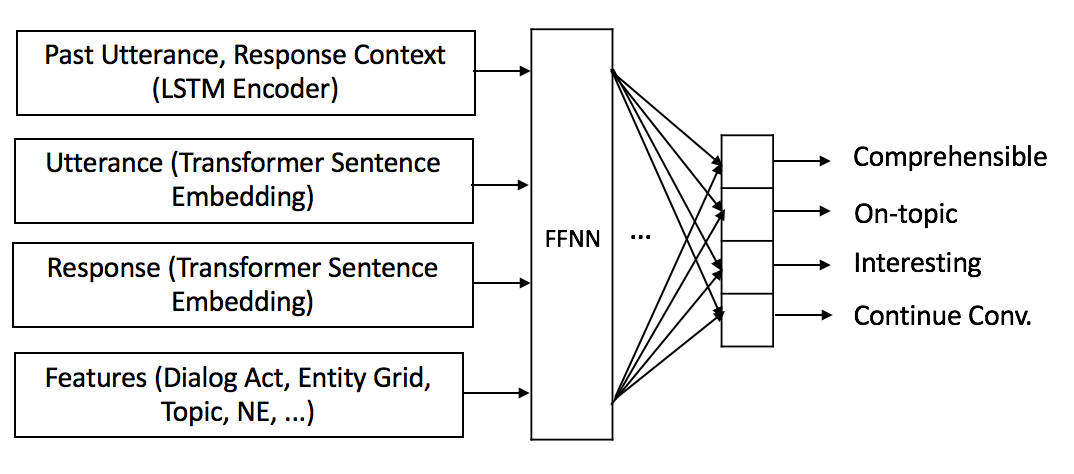}
  \caption{ Conversation Evaluators}
  \label{fig:eval_model}
\end{figure}

\subsection{Models}
Given the large number of features and their non-sequential nature, we train
four binary classifiers using feedforward neural networks~(FFNN). The input to
these models is a dialog turn. Each output layer is a softmax function
corresponding to a binary decision for each evaluation metric forming a four-dimensional vector. Each vector dimension corresponds to an evaluation metric (See
Section~\ref{subsec:Annotations}). For example, one possible reference output
would be [0,1,1,0], which corresponds to ``not comprehensible,'' ``on topic,''
``interesting,'' and ``I don't want to continue.''

We experimented with training the evaluators jointly and separately and found
that training them jointly led to better performance. We suspect this is due to
the objectives of all evaluators being closely related. We concatenate the aforementioned features as an input to a 3-layer FFNN with 256 hidden units. Figure~\ref{fig:eval_model} depicts the architecture of the conversation
evaluators.


\begin{figure}[t!]
\subfloat[{\small Baseline Response Generator (Seq2Seq with Attention)}]{
 \includegraphics[width=\linewidth]{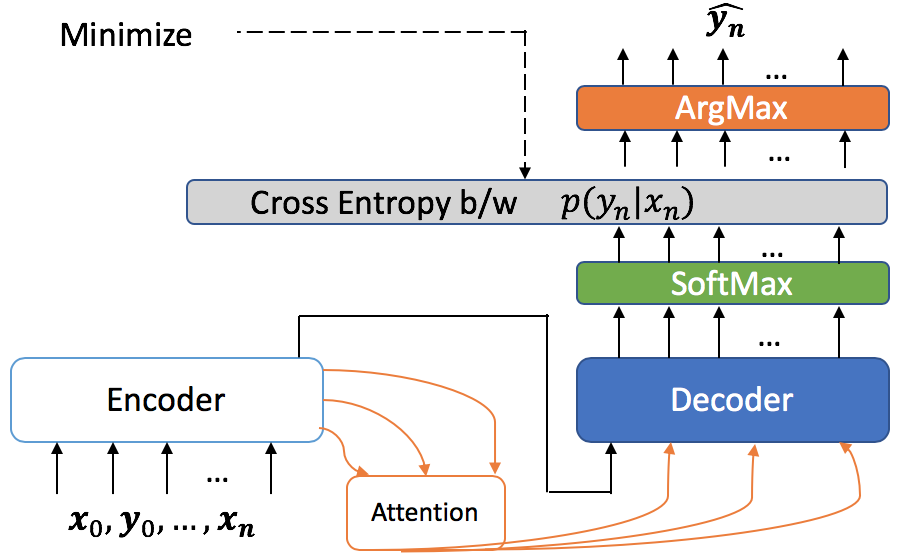}
  \label{fig:s2s_attn}
  }

 \subfloat[{\small Reranking Using Evaluators. Top $15$ candidates from beam
search are passed to the evaluators. The candidate that maximizes the reranker score is
     chosen as the output. Encoder-decoder remain unchanged.}]{
 \includegraphics[width=\linewidth]{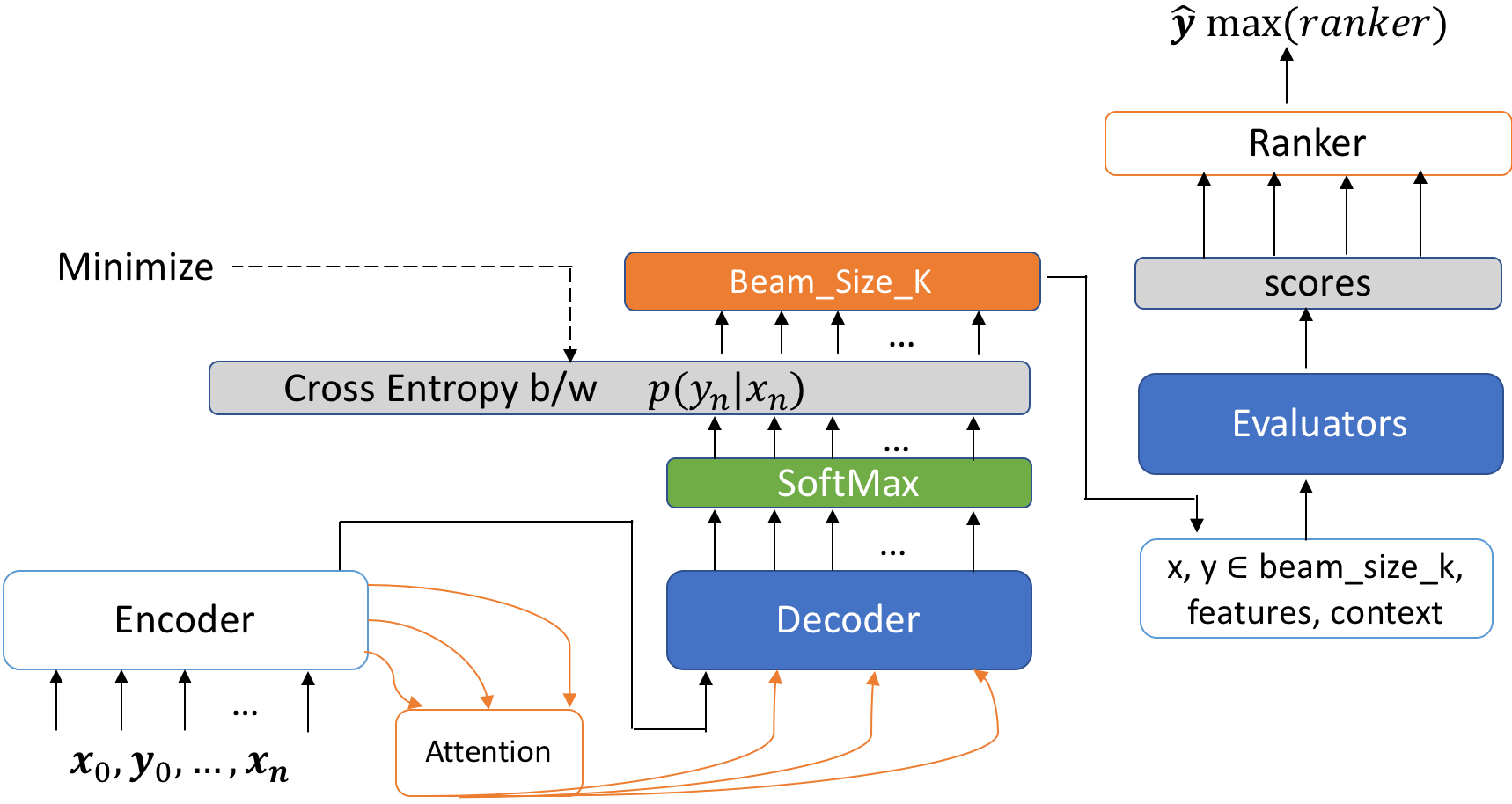}
  \label{fig:s2s_attn_rr}
  }

\subfloat[{\small Fine-tuning Using Evaluators.
We minimize cross entropy loss and maximize discriminator loss. The output of
softmax, i.e., likelihood over vocabulary for the length of output is passed to the
evaluator along with the input ($x$ and context). Evaluator generates the discriminative
score over $|V| \times len$ generator output, which is subtracted from the
loss. The updated loss is back-propagated to update encoder-decoder.}]{
  \includegraphics[width=1.0\linewidth]{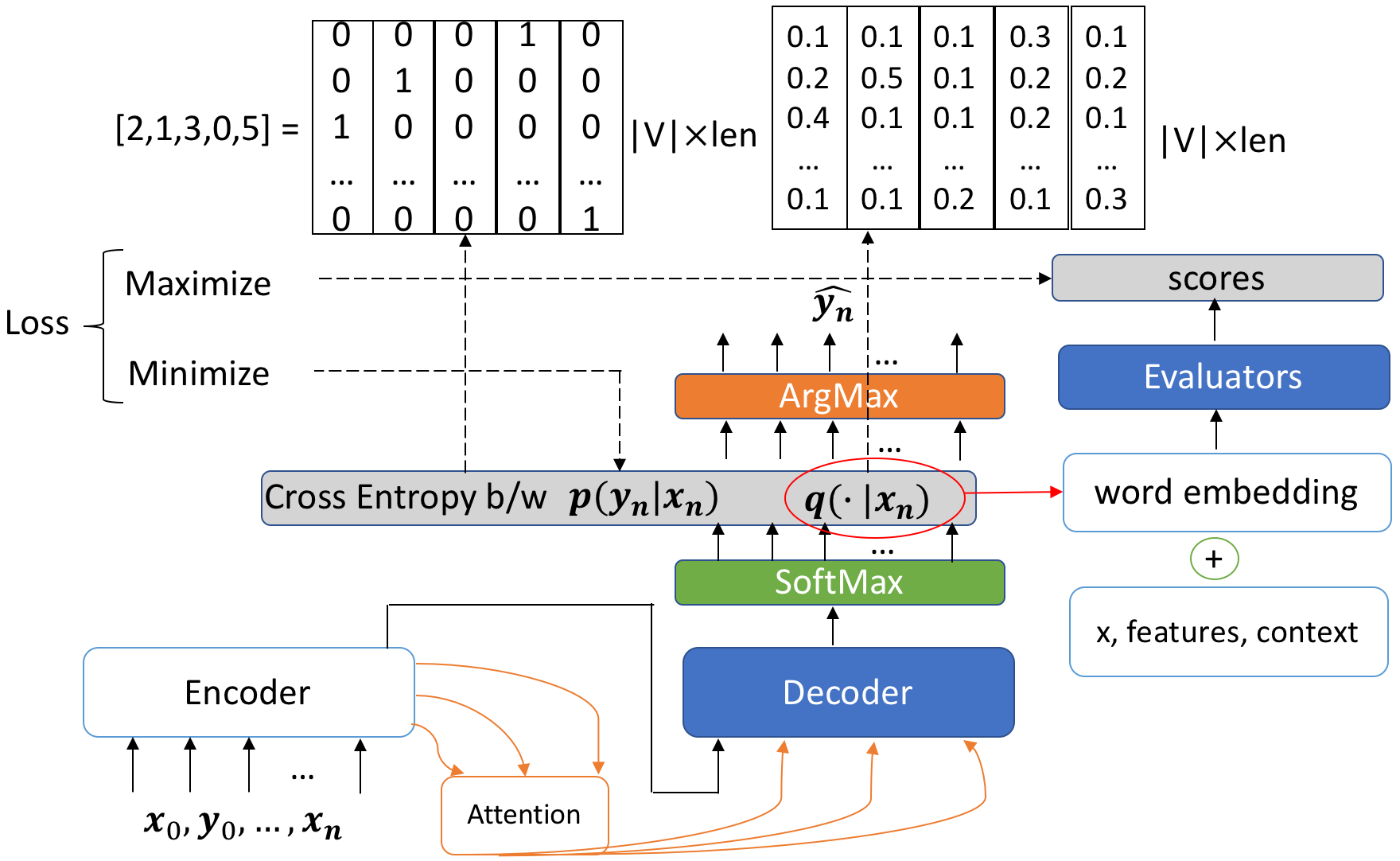}
  \label{fig:s2s_attn_ft}
  }
  \caption{Response Model Configurations. The baseline is shown at the
    top. The terms $x_{n}$ and $ y_{n}$ correspond to $n^{th}$ utterance and
    response respectively.  }
\end{figure}

\section{Response Generation System}
\label{sec:rgs}

To incorporate the explicit turn level feedback provided by the conversation
evaluators, we augment our baseline response generation system with the softmax
scores provided by the conversation evaluators. Our baseline response generation
system is described in Section~\ref{subsec:baseline}.
We then incorporate evaluators outputs using two techniques:
reranking and fine-tuning.

\subsection{Base Model (S2S)}
\label{subsec:baseline}
We extended the approach of \citet{yao2016attentional} where the authors used Luong's dot attention~\cite{luong2015effective}. 
In our experiments, the
decoder uses the same attention (Figure~\ref{fig:s2s_attn}). As we want to
observe the full impact of conversational evaluators, we do not incorporate
inverse document frequency~(IDF) or conversation topics into our
objective. Extending the objective to include these terms can be a good
direction for future work.

To make the response generation system more robust, we added user utterances and
system responses from the previous turn as context. The input to the response
generation model is previous-turn user utterance, previous-turn system response,
and current-turn user utterance concatenated sequentially. We insert a special
transition token~\cite{serban2015building} between turns. We then use a single RNN to
encode these sentences. Our word embeddings are randomly initialized and
then fine-tuned during training. We used a 1-layer Gated Recurrent Neural
network with 512 hidden units for both encoder and decoder to train the seq2seq
model and  MLE as our training objective.

\subsection{Reranking (S2S\_RR)}\label{ssec:reranker} 

In this approach, we do not update the underlying encoder-decoder model. We
maintain a beam to get 15-best candidates from the decoder. The top
candidate out of the 15 candidates is equivalent to the output of the baseline
model. Here, instead of selecting the top output, the final output response is
chosen using a reranking model.

For our reranking model, we calculate BLEU scores for each of the 15 candidate
responses against the ground truth response from the chatbot. We then sample two
responses from the $k$-best list and train a pairwise response reranker. The
response with the higher BLEU is placed in the positive class (+1) and the one
with lower BLEU is placed in the negative class (-1). We do this for all
possible candidate combinations from the 15-best responses. We use the
max-margin ranking loss to train the model. The model is a three-layered FFNN with 16 hidden units.

The input to the pairwise reranker is the softmax output of the 4 evaluators as
shown in Figure~\ref{fig:eval_model}. The input to the evaluators are described in
Section~\ref{sec:eval}. The output of the reranker is a scalar, which, if
trained right, would give a higher value for responses with higher BLEU
scores. Figure~\ref{fig:s2s_attn_rr} depicts the architecture of this model.

\subsection{Fine-tuning (S2S\_FT)}

In this approach, we use evaluators as a discriminatory loss to fine-tune the baseline encoder-decoder response generation system. We first train the baseline model and then, it is
fine-tuned using the evaluator outputs in the hope of generating more coherent
and engaging responses. One issue with MLE is that the learned models are not
optimized for the final metric (e.g., BLEU). To combat this problem, we add a
discriminatory loss in addition to the generative loss to the overall loss term
as shown in Equation~\ref{eqn:loss}. 
\begin{dmath} \label{eqn:loss}
  {loss} = \sum_{i=1}^{len} p (y_{ni}|z_n) log(q(\hat{y}_{ni}|z_n)) - \
  \lambda || {Eval} (x_{n}, q(.|z_n) ||_{1}
\end{dmath}
where $z_n = x_{n},y_{n-1},\ldots,x_{0},y_{0}$ is the conversational context where $n$ is the
context length. $q \in R^{|V| \times len}$ of the first term
corresponds to the softmax output generated by the response generation
model. The term $\hat{y}_{ni}$ refers to its corresponding decoder response at
$n_{th}$ conversation turn and $i^{th}$ word generated. In the second term, the
function $Eval$ refers to the evaluator score produced for a user utterance,
$x_n$, and decoder softmax output, $q$.

In Equation~\ref{eqn:loss}, the first term corresponds to the cross-entropy loss
from the encoder-decoder while the second term corresponds to the discriminative
loss from the evaluator.  In a standalone evaluation setting, the evaluator will
take one hot representation of the user utterance as input, i.e., the input is
$len$-tokens long which is passed through an embedding lookup layer which makes
it $\mathbb{R}^{D \times len}$ input to rest of the network where $D$ is the size of the word
embeddings.  To make the loss differentiable, instead of performing $argmax$ to
get a decoded token, we use the output of the softmax layer (distribution of
likelihood across entire vocabulary for output length, i.e., $\mathbb{R}^{|V| \times len}$)
and use this to do a weighted embedding lookup across the entire vocabulary to
get the same $\mathbb{R}^{D \times len}$ matrix as an input to rest of the
evaluator network. Our updated evaluator input becomes the following:
\begin{dmath}
  \mathbb{R}^{D \times len} = \mathbb{R}^{D \times |V|} \times \mathbb{R}^{|V| \times len}
\end{dmath}
The evaluator score is defined as the sum of softmax outputs of all 4
models.
 We keep the rest of the input~(context and features) for the evaluator
as is.


We weight the discriminator score by $\lambda$, which is a hyperparameter. We
selected $\lambda$ to be 10 using grid search to optimize for final BLEU on our development
set. Figure~\ref{fig:s2s_attn_ft} depicts the architecture of this approach. The
decoder is fine-tuned to maximize evaluator scores along while minimizing the
cross-entropy loss. The evaluator model is trained on the original annotated
corpus and parameters are frozen.

\subsection{Reranking + Fine-tuning (S2S\_RR\_FT)}
We also combined fine-tuning with reranking, where we obtained the 15 candidates
from the fine-tuned response generator and then we select the best response
using the reranker, which is trained to maximize the BLEU score.

\section{Experiments and Results}

\subsection{Conversation Evaluators}

The conversation evaluators were trained using cross-entropy
loss. We used a batch size of 128, dropout of 0.3 and Adam optimizer with a learning rate of 5e-5 for our
conversational evaluators. Sentence embeddings for user utterances and system responses are obtained
using the fast-text embeddings and  Transformer network.

Table~\ref{tab:eval_results} shows the evaluator performance compared with a
baseline with no handcrafted features.  We present precision, recall, and
f-score measures along with the accuracy.  Furthermore, since the class
distribution of the dataset is highly imbalanced, we also calculate Matthews
correlation coefficient (MCC)\cite{matthews1975comparison}, which takes into account true and false positives
and negatives. It is a balanced measure which can be used even if
the classes sizes are very different. With the proposed features we observe
significant improvement across all metrics.

We also performed a correlation study between the model predicted scores and
human annotated scores (1 to 5) on 2000 utterances. The annotators\footnote{Same
  setup as previously described} were
asked to answer a single question: ``On a scale of 1--5, how coherent and
engaging is this response given the previous conversation?'' From
Table~\ref{tab:eval_corr}, it can be observed that evaluator predicted scores
has significant correlation~(moderate to high) with the overall human evaluation
score on this subjective task (0.2 -- 0.4 Pearson correlation with turn-level
ratings). Considering the substantial individual differences in evaluating open-domain conversations, we observe that our evaluators with moderate level of correlation can be used to provide turn-level feedback for a human-chatbot conversation.

\begin{table}[]
\centering
\scalebox{0.85}{
\begin{tabular}{l|c|c}
\bf Metric & \bf {Pearson Corr} & \bf {p-value} \\
\hline
Comprehensible & 0.2 & \(<\!\!<\) 0.001 \\
On-topic & 0.4 & \(<\!\!<\) 0.001 \\
Interesting & 0.25 & \(<\!\!<\) 0.001 \\
Cont. Conversation & 0.3 & \(<\!\!<\) 0.001 \\
\end{tabular}
}
\caption { Evaluators Correlation with Turn-level Ratings} \label{tab:eval_corr} 
\end{table}

\begin{table*}[]
\centering
\scalebox{0.85}{
\begin{tabular}{l|l}
Conversation 1 & Conversation 2     \\ \hline
\begin{tabular}[c]{@{}l@{}}User: yes \\ Chatbot: is there something specific you want to discuss \\ User: fox news  \\ 
\textbf{Ground Truth}: my developers are working hard \\ to get you recent news you can say topics to hear some topics \\
\textbf{S2S(Base)}: I found \\
\textbf{S2S\_RR}: I heard about this\\
\textbf{S2S\_FT} : I heard of a man it may be \\ interested in this news mentioned \\
\textbf{S2S\_RR\_FT}:  I heard about this news talked about this \\
\end{tabular}
& \begin{tabular}[c]{@{}l@{}}User: cool \\ Chatbot: back at you ok we could talk about movies or music \\ User: movies \\ 
\textbf{Ground Truth}: shall we chat about something else \\ I love talking about music and movies \\
\textbf{S2S(Base)}: so do you \\
\textbf{S2S\_RR}:  who would you want to talk \\
\textbf{S2S\_FT} : what actor love most \\
\textbf{S2S\_RR\_FT}:  what actor  \\
\end{tabular} 
\end{tabular}
}
\caption{\small Two randomly selected qualitative examples of responses} \label{tab:generator_examples}
\end{table*}

\subsection{Response Generation}

We first trained the baseline model (S2S) on the conversational data set (4M utterance-response pairs from the competition. Section~\ref{sec:data}). The data
were split into 80\% training, 10\% development, and 10\% test sets. The baseline model was trained using Adam with learning rate of 1e-4 and batch size of 256 until the development loss converges. The vocabulary of 30K most frequent words were used. And the reranker was trained using the 20K number of beam outputs from the baseline model on the development set. Adam with learning rate of 1e-4 and batch size of 16 was used for the fine-tuning (S2S\_FT).

Table~\ref{tab:generator_ap_results} shows the performance
comparison of different generation models (Section~\ref{sec:rgs}) on the Alexa Prize conversational data set.
We observed that reranking $n$-best responses using the evaluator-based reranker~(S2S\_RR)
provides nearly 100\% improvement in BLEU-4 scores.
\begin{table}[]
\scalebox{0.82}{
\begin{tabular}{l|r|r|r}
\bf Metric                        & BLEU-4         & ROUGE-2  & Distinct-2                          \\ \hline
\bf {S2S (Base)}        & 5.9          & 5.1 &   0.011                          \\
\bf {S2S\_RR}     & 11.6(+97\%)  & 6.3(+24\%)   & \textbf{0.017(+54\%)}                   \\
\bf {S2S\_FT}      & 6.2(+5\%)   & 5.3(+4\%) & 0.011(-1\%)                      \\
\bf {S2S\_RR\_FT}  & \textbf{12.2(+107\%)} & \textbf{6.8(+33\%)} & 0.017(+53\%) \\
\end{tabular}
}
\caption { Generator performance on automatic metrics.} \label{tab:generator_ap_results} 
\end{table}

Fine-tuning the generator by adding  evaluator loss
(S2S\_FT) does improve the performance but the gains are smaller compared to reranking. We suspect that this is due to the reranker directly optimizing
for BLEU. However, using a fine-tuned model and then reranking (S2S\_RR\_FT)
complements each other and gives the best performance overall. Furthermore, we
observe that even though the reranker is trained to maximize the BLEU scores,
reranking shows significant gains in ROUGE scores as well. We also measured different systems performance
using Distinct-2~\cite{li2016diversity}, which is the number of unique
length-normalized bigrams in responses. The metric can be a surrogate for measuring diverse outputs. We see that our generators using reranking approaches improve on this metric as well. Table~\ref{tab:generator_examples} also shows 2 sampled responses from different models.

To further analyze the impact of reranker trained to optimize on BLEU score, we
trained a baseline response generation system on a Reddit data
set\footnote{We use a publicly available data \cite{reddit}.},
which comprises of 9 million comments and corresponding response comments. All the hyperparameter setting followed the setting of training on the Alexa Prize conversational dataset. 
 
We trained a new reranker for the Reddit data using the evaluator
scores obtained from the models proposed in Section~\ref{sec:eval}. We show in
Table~\ref{tab:generator_reddit_results} that even though the
evaluators are trained on a different data set, the reranker 
learns to select better responses nearly doubling the BLEU
scores as well as improving on the Distinct-2 score. Thus, the evaluator generalizes in selecting more coherent and engaging
responses in human-human interactions as well as human-computer interactions. As fine-tuning
the evaluator is computationally expensive, we did not fine-tune it on the
Reddit dataset.

The closest baseline that used BLEU scores for evaluation in open-domain
setting is from \citet{li2015diversity} where they trained the models on
Twitter data using Maximum Mutual Information (MMI) as the objective
function. They obtained a BLEU score of 5.2 in their best setting on Twitter
data (average length 23 chars), which is relatively less complex than Reddit
(average
length 75 chars).


\begin{table}[]
\centering
\scalebox{0.85}{
\begin{tabular}{l|c|c|c|c}
\bf Metric & \bf {S2S(Base)} & \bf {S2S\_RR} \\ \hline
BLEU-4 & 3.9 & 7.9 (+103\%) \\
ROUGE-2 & 0.6 & 0.8 (+33\%) \\
Distinct-2 & 0.0047 & 0.0086 (+82\%) \\
 \end{tabular}}
 \caption {Response Generator on Reddit Conversations. Due to the size of
the dataset we could not fine tune these models.}
\label{tab:generator_reddit_results} 
\end{table}

\begin{table}[]
  \centering    
  \scalebox{0.85}{

    \begin{tabular}{l|c|c}
      \bf Metric                & Coherence    & Engagement                          \\ \hline
      \bf {S2S(Base)}  & 2.34                    & 1.80                            \\
      \bf {S2S\_RR}     & 2.42                    & 2.16                  \\
      \bf {S2S\_FT}     & 2.36                    & 1.87                  \\
      \bf {S2S\_RR\_FT} & \bf 2.55 & \bf 2.31   \\
    \end{tabular}
  }
  \caption { Mean ratings for Qualitative and Human Evaluation of Response Generators} \label{tab:generator_human_eval}
\end{table}

\subsection{Human Evaluation}
As noted earlier, automatic evaluation metrics may not be the best way to
measure chatbot response generation performance. Therefore, we performed human
evaluation of our models. We asked annotators to provide ratings on the system
responses from the models we evaluated, i.e., baseline model, S2S\_RR, S2S\_FT,
and S2S\_RR\_FT.
A rating was obtained on
two metrics: coherence and engagement. Coherence measures how much the response is comprehensible and relevant to a user's request and engagement shows interestingness of the response (\citet{venkatesh2018evaluating}). We asked the annotators to
provide the rating based on a scale of 1--5, with 5 being the best. We had four
annotators rate 250 interactions. Table~\ref{tab:generator_human_eval}
shows the performance of the models on the proposed metrics. Our
inter-annotator agreement is 0.42 on Cohen's Kappa Coefficient, which implies
 moderate agreement. We believe this is because the task is
relatively subjective and the conversations were performed in the challenging
open-domain setting. The S2S\_RR\_FT model provides the best performance
across all the metrics, followed by S2S\_RR, followed by S2S\_FT.

\section{Conclusion}
Human annotations for conversations show significant variance, but it is still
possible to train models which can extract meaningful signal from the
human assessment of the conversations. We show that these models
can provide useful turn-level guidance to response generation models. We design
a system using various features and context encoders to provide
turn-level feedback in a conversational dialog. Our feedback is interpretable on
2 major axes of conversational quality: engagement and coherence. We also plan to provide similar evaluators to the university teams participating in the Alexa Prize competition. 
 To show that such feedback is useful in building better conversational response systems, we propose 2
ways to incorporate this feedback, both of
which help improve on the baselines. Combining both techniques results in the best performance. We view this work as complementary to other recent work
in improving dialog systems such as~\citet{li2015diversity}
and~\citet{shao2017generating}. While such open-domain systems are still in
their infancy, we view the framework presented in this paper to be an important
step towards building end-to-end coherent and engaging chatbots.

\bibliography{refs}
\bibliographystyle{acl_natbib}

\end{document}